\documentclass[lettersize,journal]{IEEEtran}
\usepackage{amsmath,amsfonts}
\usepackage{algorithmic}
\usepackage{algorithm}
\usepackage{array}
\usepackage[caption=false,font=normalsize,labelfont=sf,textfont=sf]{subfig}
\usepackage{textcomp}
\usepackage{stfloats}
\usepackage{url}
\usepackage{verbatim}
\usepackage{graphicx}
\usepackage[small]{caption}
\usepackage{cite}
\usepackage{hyperref}
\usepackage{booktabs}  
\usepackage{multirow}   
\usepackage[table]{xcolor}
\usepackage{tabularx}
\usepackage{siunitx}
\usepackage{tikz}
\usetikzlibrary{calc}

\usepackage{colortbl}
\usepackage{hhline}   
\usepackage{makecell} 
\usepackage[numbers,sort&compress]{natbib}

\newcommand{\ie}{\textit{i}.\textit{e}., }
\newcommand{\eg}{\textit{e.g., }}
\newcommand{\etc}{\textit{etc.}}
\newcommand{\vs}{\textit{v.s.}}

\definecolor{MyDarkRed}{RGB}{192, 0, 0}
\definecolor{MyDarkGreen}{RGB}{84, 130, 53}

\begin{document}

\title{Structure-aware Prompt Adaptation from Seen to Unseen for Open-Vocabulary Compositional Zero-Shot Learning}

\author{
    Yihang~Duan$^{1}$, 
    Jiong~Wang$^{2}$, 
    Pengpeng~Zeng$^{1}$, 
    Ji~Zhang$^{3}$, 
    Lei~Zhao$^{1}$, 
    Chong~Wang$^{2}$, \\
    Jingkuan~Song$^{4}$, 
    Lianli~Gao$^{1}$ \\
    \vspace{0.5em}
    $^{1}$University of Electronic Science and Technology of China, \\
    $^{2}$Ningbo University, 
    $^{3}$Southwest Jiaotong University, 
    $^{4}$Tongji University \\
}

\maketitle

\begin{abstract}
The goal of Open-Vocabulary Compositional Zero-Shot Learning (OV-CZSL) is to recognize attribute-object compositions in the open-vocabulary setting, where compositions of both seen and unseen attributes and objects are evaluated.
Recently, prompt tuning methods have demonstrated strong generalization capabilities in the closed setting, where only compositions of seen attributes and objects are evaluated, \ie Compositional Zero-Shot Learning (CZSL). 
However, directly applying these methods to OV-CZSL may not be sufficient to generalize to unseen attributes, objects and their compositions, as it is limited to seen attributes and objects. 
Normally, when faced with unseen concepts, humans adopt analogies with seen concepts that have the similar semantics thereby inferring their meaning (\eg ``wet'' and ``damp'', ``shirt'' and ``jacket''). In this paper, we experimentally show that the distribution of semantically related attributes or objects tends to form consistent local structures in the embedding space. 
Based on the above structures, we propose \textbf{S}tructure-aware \textbf{P}rompt \textbf{A}daptation (\textbf{SPA}) method, which enables models to generalize from seen to unseen attributes and objects.
Specifically, in the training stage, we design a Structure-aware Consistency Loss (SCL) that encourages the local structure's consistency of seen attributes and objects in each iteration. 
In the inference stage, we devise a Structure-guided Adaptation Strategy (SAS) that adaptively aligns the structures of unseen attributes and objects with those of trained seen attributes and objects with similar semantics. Notably, SPA is a plug-and-play method that can be seamlessly integrated into existing CZSL prompt tuning methods.
Extensive experiments on the OV-CZSL benchmarks prove the effectiveness of SPA, specifically its ability to achieve competitive performance in the closed setting while significantly improving results in the open-vocabulary setting.
The code is available at: \url{https://github.com/ZHlo-404/SPA}.
\end{abstract}


\section{Introduction} \label{intro}

\begin{figure}[t]
    \centering
    \includegraphics[width=\linewidth]{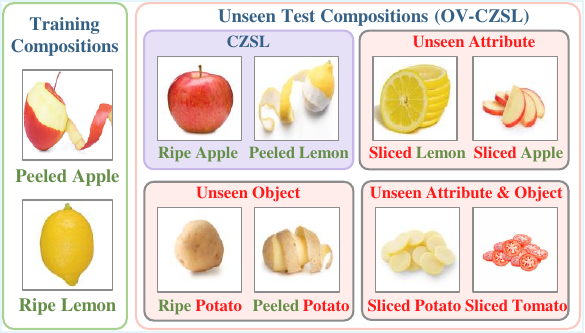}
    \caption{Illustration of Compositional Zero-Shot Learning (CZSL) and its extended version Open-Vocabulary Compositional Zero-Shot Learning (OV-CZSL).}
    \label{fig:czsl_ovczsl_setting} 
    \vspace{-1.5em}
\end{figure}

Compositional Zero-Shot Learning (CZSL) aims to recognize unseen compositions made of attributes and objects that have appeared within seen compositions during training. 
While existing CZSL methods~\cite{DBLP:conf/iclr/NayakYB23,wang2023hierarchical,DBLP:conf/cvpr/LuGLG23,huang2024troika,min2024adaptive} achieve promising results, they rely on a closed-set assumption where all attributes and objects are predefined during training.
This constraint limits their applicability to real-world scenarios, where new attributes and objects frequently emerge.
To overcome this, Open-Vocabulary CZSL (OV-CZSL)~\cite{saini2024beyond} extends the task to an open-world setting, enabling models to recognize novel attribute–object compositions beyond the training vocabulary.
As illustrated in Fig.~\ref{fig:czsl_ovczsl_setting}, while CZSL considers only compositions of seen attributes and objects during training, OV-CZSL extends this setting to include three additional types: seen attributes with unseen objects, unseen attributes with seen objects, and compositions involving both unseen attributes and objects.

Recently, prompt tuning with pre-trained vision-language models(\ie CLIP) has demonstrated remarkable generalization ability in CZSL, effectively modeling attribute-object compositions.
Among these, CSP\cite{DBLP:conf/iclr/NayakYB23} stands out as an early CLIP-based approach that treats attributes and objects as learnable textual tokens and integrates them into prompt templates (\ie "a photo of [attribute] [object]") to enhance the compositional representation learning. Building on this foundation, follow-up works such as DFSP\cite{DBLP:conf/cvpr/LuGLG23} and Troika~\cite{huang2024troika} introduce elaborately designed adaptation and interaction mechanisms to further improve CZSL performance. 
However, in contrast to CZSL, the main challenge of OV-CZSL is enabling models to effectively generalize from seen to unseen, particularly when handling compositions that involve novel attributes, objects, or both. 
To tackle this challenge, Saini et al.~\cite{saini2024beyond} propose Neighborhood Expansion Loss (NEL), which incorporates external semantic concepts to enhance generalization. 
However, this method relies on BERT for textual encoding and an ImageNet-pretrained visual encoder, whose representation capabilities are relatively weak and less effective at capturing fine-grained attribute-object interactions. 
This raises an intuitive question: \textit{Can we leverage the powerful representations of CLIP to better address the challenges of OV-CZSL?}

\begin{figure}[t]
    \centering
    \includegraphics[width=\linewidth]{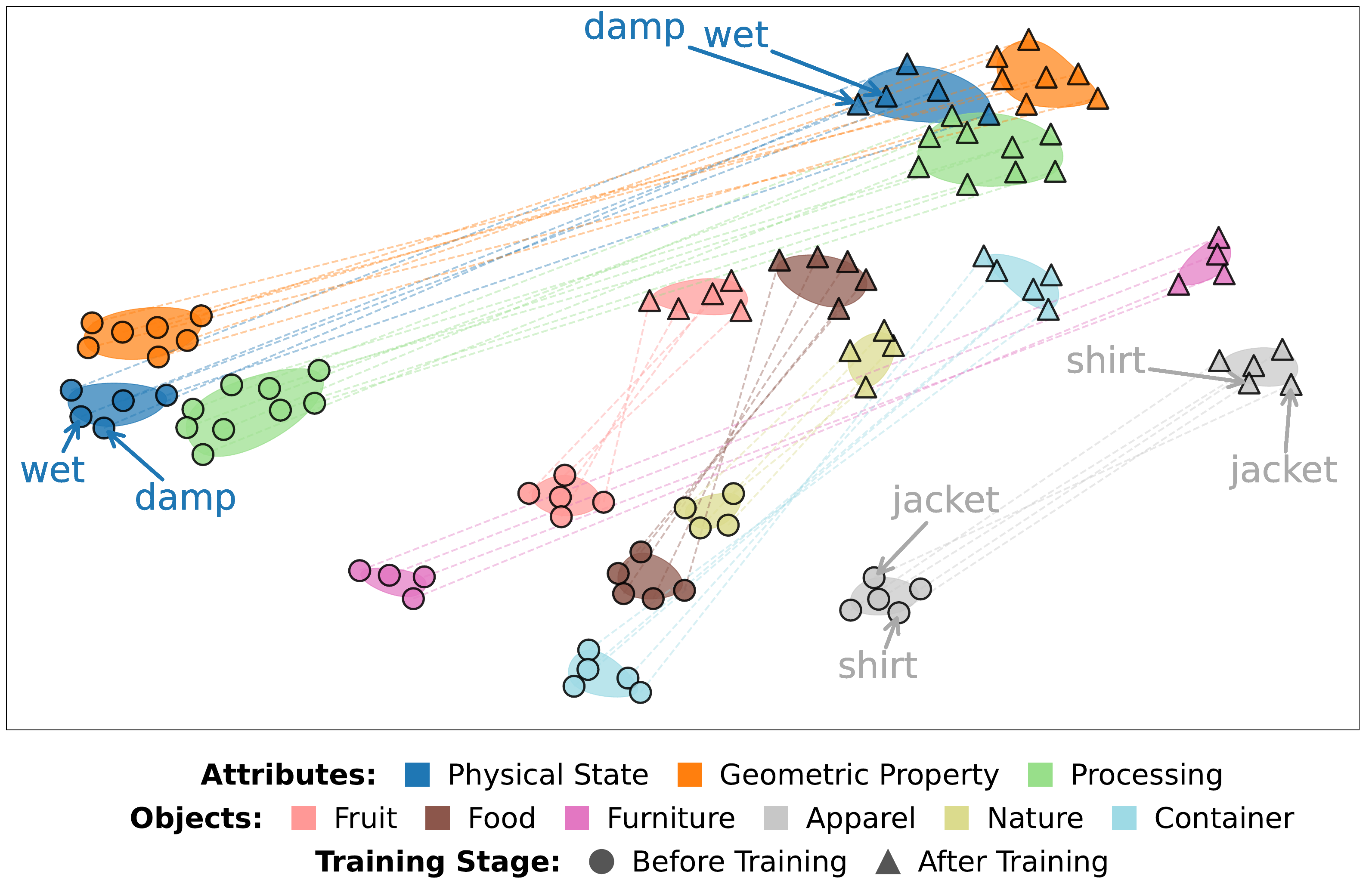}
    \caption{\textbf{Our core motivation}: Semantically related primitives (e.g., "wet" and "damp") form consistent local structures in the embedding space. This structural prior is preserved through training and provides a basis for generalizing from seen to unseen concepts.}
    \label{fig:embedding_visualization} 
    \vspace{-1.5em}
\end{figure}

With this question in mind,  we empirically conduct a confirmatory experiment applying CLIP-based prompt tuning methods to the OV-CZSL task.
As shown in Tab.~\ref{tab:mit&cgqa}, methods like CSP significantly outperform non-CLIP baselines (\ie 26.82 \vs 10.94 HM on MIT-States), highlighting CLIP’s strong representation power in this setting.
However, we observe that the performance gains primarily remain within the seen domain, and the generalization to unseen attributes and objects is still limited.
To further address this limitation, we turn to human cognitive behavior. Humans are often able to infer unfamiliar concepts by associating them with semantically similar known concepts (\eg understanding “damp” through “wet”, or relating “jacket” to “shirt”). Inspired by this, we choose a representative CLIP-based model, CSP~\cite{DBLP:conf/iclr/NayakYB23}, and visualize the embedding space using t-SNE. As shown in Fig.~\ref{fig:embedding_visualization}, we find that semantically related attributes and objects (\eg “shirt” and “jacket” under the "Apparel" category) tend to cluster together and form local structures in CLIP’s embedding space. Interestingly, this structural coherence is largely preserved before and after training, indicating that these local relationships could serve as a valuable prior to guide the learning of unseen concepts. Indeed, the effectiveness of relying on these local structures is substantiated by recent advances in related areas like zero-shot classification~\cite{kalantidis2024label} and robust learning~\cite{chang2023csot,zhangji2025reliable}, confirming the reliability of CLIP's local structure for knowledge transfer.

Based on the above observations, we propose Structure-aware Prompt Adaptation (SPA), an effective framework designed to improve compositional generalization in the open-vocabulary setting. SPA aims to leverage the inherent local structures in CLIP’s embedding space to facilitate knowledge transfer from seen to unseen attribute-object compositions.
During the training stage, we introduce a Structure-aware Consistency Loss (SCL), which encourages the preservation of local structural relationships among seen attributes and objects between their original CLIP-based embeddings and their updated embeddings after prompt tuning. This consistency constraint helps maintain the semantic neighborhood formed by CLIP and ensures that fine-tuning does not distort the meaningful relational structure.
At inference time, we propose a Structure-guided Adaptation Strategy (SAS) that dynamically adjusts the embeddings of unseen attributes and objects by aligning their structures with the local structures of semantically similar seen attributes and objects.
This alignment allows the model to better integrate unseen concepts into the learned compositional space, thereby enhancing recognition performance on unseen compositions.
Notably, SPA is a plug-and-play method that can be seamlessly integrated into existing prompt tuning methods.

Our main contributions are as follows:
\begin{itemize}
    \item {We present a pioneering exploration of CLIP-based prompt tuning for Open-Vocabulary Compositional Zero-Shot Learning (OV-CZSL), demonstrating its strong potential in this challenging setting.}
    \item We propose Structure-aware Prompt Adaptation (SPA), which utilizes local structural consistency to generalize from seen to unseen attributes and objects, combining Structure-aware Consistency Loss (SCL) to preserve local coherence in seen concepts and Structure-guided Adaptation Strategy(SAS) to align unseen concepts with the learned structure, improving recognition performance on unseen compositions.
    \item Extensive experiments on multiple OV-CZSL benchmarks demonstrate that our proposed SPA is both effective and flexible, without introducing significant computational overhead.
\end{itemize}

\section{Related Works}

\subsection{Compositional Zero-shot learning (CZSL)}
Compositional Zero-Shot Learning (CZSL) aims to recognize novel attribute-object combinations from previously seen components (\eg inferring "peeled apple" from seen "ripe apple" and "peeled lemon"). Early studies on CZSL can be broadly categorized into two directions, namely joint and disentangled representations. 
Representative works~\cite{naeem2021learning,nagarajan2018attributes,anwaar2022leveraging,zhangji2023channel} learn unified embeddings for attribute–object pairs to capture compositional semantics within a shared space.
In contrast, subsequent studies~\cite{li2022siamese,atzmon2020causal,huynh2020compositional,karthik2022kg,yang2022decomposable} disentangle attribute and object representations, enabling independent modeling and improving generalization to unseen compositions.
Recently, the focus has shifted to leveraging Vision-Language Pre-trained (VLP) models like CLIP~\cite{radford2021learning}. For instance, CSP~\cite{DBLP:conf/iclr/NayakYB23} uses trainable tokens for attributes and objects in prompts, while DFSP~\cite{lu2023decomposed} employs a decomposed fusion module to enhance fine-grained interactions.
However, traditional CZSL struggles with entirely unseen attributes or objects. To address this limitation, Saini et al.~\cite{saini2024beyond} introduce the Open-Vocabulary CZSL (OV-CZSL) setting, but their method is incompatible with modern prompt-tuning techniques. 
{In this work, we present a pioneering exploration of CLIP-based prompt tuning for OV-CZSL, proposing a plug-and-play method to bridge this gap and significantly enhance the generalization of existing baselines~\cite{DBLP:conf/iclr/NayakYB23,wang2023hierarchical,DBLP:conf/cvpr/LuGLG23,huang2024troika} in open-vocabulary scenarios.}

\subsection{Open Vocabulary Learning}
Open vocabulary learning aims to enable models to comprehend arbitrary text-described concepts by semantically aligning visual and linguistic modalities, thereby overcoming the limitations of predefined category labels. Pioneering works like CLIP~\cite{radford2021learning} and ALIGN~\cite{jia2021scaling} established this paradigm through large-scale image-text contrastive learning, projecting images and text into a shared embedding space to achieve zero-shot classification. Building on this foundation, the OV approach has been extended to more complex vision tasks, including object detection~\cite{DBLP:conf/iclr/GuLKC22,zhangji2023global}, semantic segmentation~\cite{DBLP:conf/iclr/LiWBKR22}, and video tracking~\cite{li2023ovtrack}. 
{Despite these advancements, practical scenarios necessitate a critical shift from identifying isolated atomic concepts toward reasoning about their intricate compositional interactions. This necessity underpins the emerging task of Open-Vocabulary Compositional Zero-Shot Learning (OV-CZSL)~\cite{saini2024beyond}, which moves beyond primitive-level recognition to address the more challenging goal of comprehending novel attribute-object combinations where both components may be entirely unseen.}

\subsection{Prompt Learning}
Prompt learning adapts pre-trained models to downstream tasks by optimizing input prompts\cite{zhangjiwang2025dual,zhangji2025closer}. Originating from NLP, this technique has evolved from using rigid, manually crafted templates to learnable prompts that automatically adjust task-specific representations while keeping the backbone model frozen.
Building on this success, the paradigm extends naturally to Vision-Language Models (VLMs). CoOp~\cite{DBLP:journals/ijcv/ZhouYLL22} first introduced learnable context vectors to replace static text prompts in CLIP~\cite{radford2021learning}, enabling end-to-end adaptation without fine-tuning the encoders. However, CoOp employs a single prompt, which makes it difficult to disentangle compositional semantics. To overcome this limitation, CSP~\cite{DBLP:conf/iclr/NayakYB23} proposes primitive-aware prompting, decomposing prompts into separate, learnable embeddings for attributes and objects (e.g., “a photo of [attribute] [object]”). This design allows independent optimization of primitives and enhances zero-shot compositional reasoning. Subsequent methods like DFSP~\cite{DBLP:conf/cvpr/LuGLG23} further refine this idea with improved disentanglement strategies. 
Despite their success, methods based on the CSP paradigm~\cite{DBLP:conf/iclr/NayakYB23,wang2023hierarchical,DBLP:conf/cvpr/LuGLG23,huang2024troika} are inherently limited in open-vocabulary settings, as they cannot model unseen attributes and objects. To overcome this limitation, we propose a plug-and-play method that adapts these CZSL approaches to the OV-CZSL task, significantly improving their generalization to novel concepts.

\section{Methods}
\begin{figure*}[]
    \centering
    \includegraphics[width=0.9\linewidth]{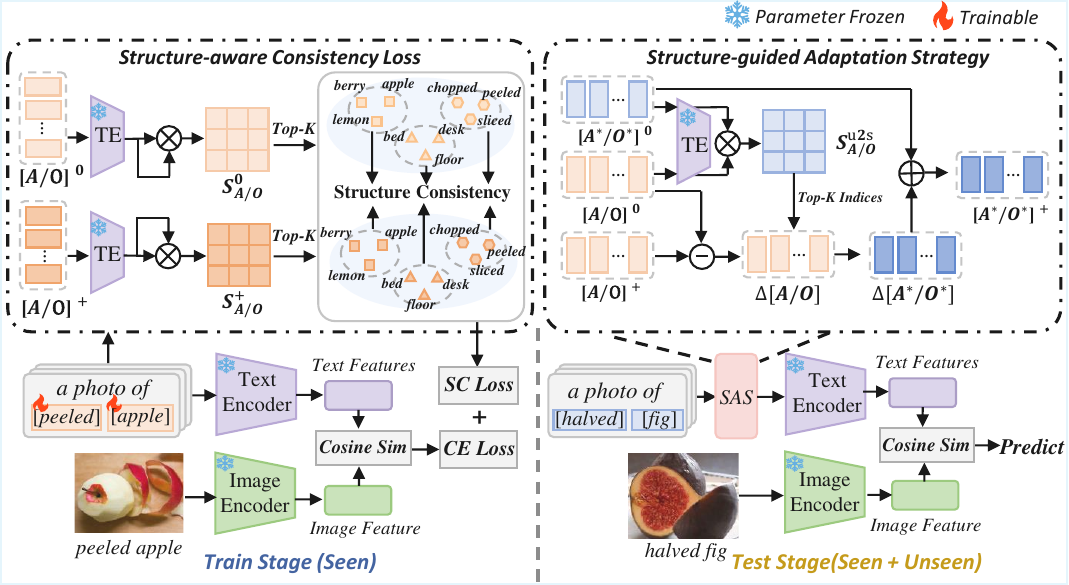}
    \caption{\textbf{Overview of the proposed Structure-aware Prompt Adaptation (SPA).} It comprises two main components: (1) Structure-aware Consistency Loss (SCL) applied during training to preserve the structural distribution of seen primitives in the CLIP embedding space, and (2) Structure-guided Adaptation Strategy dynamically adjusts the representations of unseen primitives by using their semantic proximity to seen ones at test time. SPA operates as a plugin that seamlessly integrates with existing CLIP-based prompt tuning methods.}
    \label{framework}
    \vspace{-1.5em}
\end{figure*}

In this section, we present Structure-aware Prompt Adaptation (SPA), a plug-and-play module designed to enhance CLIP-based prompt tuning methods for open-vocabulary compositional zero-shot learning (OV-CZSL). Specifically, we start with an overview of essential preliminaries, including the task formulation of OV-CZSL and the CLIP-based prompt tuning paradigm in Sec.~\ref{3.1}. Next, we illustrate the details of SPA in Sec.~\ref{3.2}, containing two novel components, \ie Structure-aware Consistency Loss (SCL) and Structure-guided Adaptation Strategy (SAS).
Finally, the training objectives and inference procedure are described  in Sec.~\ref{3.3}. The overall architecture can be seen in Fig.\ref{framework}.

\subsection{Preliminary} \label{3.1}
\noindent\textbf{Task Formulation.} 
The goal of Open-Vocabulary Compositional Zero-Shot Learning (OV-CZSL) is to recognize images described by attribute-object compositions, where either the attribute, the object, or both may be unseen during training. 
Formally, let \( A \) and \( O \) denote the sets of seen attributes and objects, respectively, and \( A^* \), \( O^* \) represent unseen attributes and objects. 
Each image is annotated with a pair \( (a, o) \), where \( a \in A \cup A^* \), \( o \in O \cup O^* \). 
Notably, all pairs in the compositional space are valid pairs, partitioned into seen compositions \( Y^s \subseteq AO \). and unseen compositions \( Y^u \), the latter comprising four distinct cases:

\begin{itemize}
    \item \( (AO)^* \): novel compositions of seen attributes and seen objects excluded from training
    \item \( AO^* \): compositions of seen attributes with previously unseen objects
    \item \( A^*O \): compositions of previously unseen attributes with seen objects
    \item \( A^*O^* \): compositions involving both unseen attributes and unseen objects
\end{itemize}

Given training data $\mathcal{D}^{tr} = \{(x_i, y_i)\}$ where $y_i \in Y^s$, the objective is to learn a model $\mathcal{M}: x \rightarrow y \in Y = Y^{s} \bigcup Y^{u}$ that generalizes to all types of compositions under the strict separation constraint \( Y^s \cap Y^u = \emptyset \). This formulation extends conventional CZSL (where $A^* = \emptyset, O^* = \emptyset$) by allowing compositions to include previously unseen attributes and objects, thereby unifying compositional generalization with open-vocabulary recognition.

\noindent\textbf{Revisiting the CLIP-based Prompt Tuning Paradigm.}
Our work builds upon the CLIP-based prompt tuning paradigm, which we briefly revisit here.
CLIP~\cite{radford2021learning}, a vision-language model pre-trained on large-scale image-text pairs, has demonstrated strong performance across a range of tasks, such as image classification\cite{abdelfattah2023cdul,peng2023sgva} and object detection\cite{zhong2022regionclip,gu2021open}. 
Recent progress in CZSL has been driven by prompt tuning with CLIP. In this framework, the vision and text encoders are frozen, while the prompts for attributes and objects are learned to enable flexible recomposition and recognition of attribute–object pairs.

Given an input image $x_i$, its visual embedding is extracted via the frozen CLIP vision encoder:
\begin{equation}
    v_i = f_{\text{vis}}(x_i) \in \mathbb{R}^d,
\end{equation}
where $f_{\text{vis}}(\cdot)$ denotes the frozen vision encoder and $d$ is the embedding dimension.

Each candidate composition label $y_j = (a_j, o_j) \in Y = Y^s \cup Y^u$ is converted into a natural language prompt using a predefined template, such as \textit{"a photo of a [attribute] [object]"}, where \textit{[attribute]} and \textit{[object]} are learnable tokens. The CLIP text encoder then encodes each prompt to a corresponding text embedding:
\begin{equation}
    t_{y_j} = f_{\text{text}}(\text{P}_j) \in \mathbb{R}^d,
\end{equation}
where $\text{P}_j$ is the prompt for composition $y_j$, and $f_{\text{text}}(\cdot)$ is the frozen CLIP text encoder.

The alignment between the image and each candidate composition is measured by cosine similarity:
\begin{equation}
    S_{i,j} = \cos(v_i, t_{y_j}) = \frac{v_i^\top t_{y_j}}{\|v_i\| \cdot \|t_{y_j}\|}.
\end{equation}

The prediction probability over all candidate compositions is then given by a temperature-scaled softmax function:
\begin{equation}
    P(y_j | x_i) = \frac{\exp(S_{i,j} / \tau)}{\sum_{k=1}^{|Y|} \exp(S_{i,k} / \tau)},
\end{equation}
where $\tau$ is a temperature parameter controlling the sharpness of the distribution.

Finally, the model is trained using cross-entropy loss over the seen composition set $Y^s$:
\begin{equation}
    \mathcal{L}_{\text{CE}} = - \frac{1}{N} \sum_{i=1}^N \log P(y_i | x_i),
\end{equation}
where $x_i$ is an image labeled with $y_i \in Y^s$, and $N$ is the number of training samples.

\subsection{Structure-aware Prompt Adaptation} \label{3.2}
Building on the insight that semantically similar primitives (attributes and objects) exhibit local structure consistency in their embedding space, we propose Structure-aware Prompt Adaptation (SPA) to preserve such structure and enable generalization from seen to unseen compositions for OV-CZSL task.


\noindent\textbf{Structure-aware Consistency Loss.}
As discussed in Sec.~\ref{intro}, semantically similar primitives (attributes or objects) form local structures in the embedding space. Through large-scale pretraining on text-image alignment tasks, CLIP learns rich semantic prior knowledge and maintains well-structured local neighborhoods (\eg the "wet" and "damp" attributes are closely embedded). However, direct fine-tuning on downstream datasets may overfit to the training distribution and distort these original structures. To address this, we introduce a Structure-aware Consistency Loss (SCL) that regularizes the local structural relationships of seen primitives during training.

We first extract text embeddings of attributes and objects using CLIP’s text encoder. Each attribute or object is represented as a prompt-based textual description, where the learnable parameters $\theta_p$ ($p \in \{a, o\}$) are optimized during fine-tuning. Specifically, we construct the following prompts:

\begin{equation}
    \begin{aligned}
        \mathbf{P}_p^{(0)} &= [w_0, \dots, w_m, \theta_p^{(0)}], \quad 
        \mathbf{P}_p^{(+)} = [w_0, \dots, w_m, \theta_p^{(+)}]
    \end{aligned}
\end{equation}
where [$w_0, \dots, w_m$] are fixed tokens representing the context, and $\theta_p$ denotes the learnable parameters for either attributes ($\theta_a$) or objects ($\theta_o$). The superscripts $(0)$ and $(+)$ indicate the embeddings from the pretrained CLIP and the embeddings being optimized during fine-tuning, respectively. 
By feeding these prompts into the frozen text encoder $f_{\text{text}}$, we obtain the corresponding embeddings:

\begin{equation}
    \begin{aligned}
        \mathbf{t}_p^{(0)} &= f_{\text{text}}(\mathbf{P}_p^{(0)}), \quad 
        \mathbf{t}_p^{(+)} = f_{\text{text}}(\mathbf{P}_p^{(+)})
    \end{aligned}
\end{equation}
where $p \in \{a, o\}$ denotes either attribute or object.
To quantify the structural relationships among attributes and among objects, we compute pairwise cosine similarity matrices before and after fine-tuning:

\begin{equation}
    \begin{aligned}
        \mathbf{S}_p^{(0)} &= \frac{\mathbf{t}_p^{(0)} (\mathbf{t}_p^{(0)})^\top}{\|\mathbf{t}_p^{(0)}\|^2}, \quad 
        \mathbf{S}_p^{(+)} = \frac{\mathbf{t}_p^{(+)} (\mathbf{t}_p^{(+)})^\top}{\|\mathbf{t}_p^{(+)}\|^2}
    \end{aligned}
\end{equation}

To capture the local structural relationships, we define the neighborhood of each attribute or object based on its Top-K most similar neighbors in the initial similarity matrix $\mathbf{S}_p^{(0)}$. This neighborhood is treated as the primitive’s local structure and remains fixed during training:

\begin{equation}
    \mathcal{I}_p = \text{TopK}(\mathbf{S}_p^{(0)}), \quad
    \mathbf{S}_p^{(0,\mathcal{N})} = \mathbf{S}_p^{(0)}[\mathcal{I}_p], \quad
    \mathbf{S}_p^{(+,\mathcal{N})} = \mathbf{S}_p^{(+)}[\mathcal{I}_p]
\end{equation}
Here, $\mathcal{I}_p$ denotes the indices of the Top-K neighbors in the initial space, which capture the local structure of attributes or objects. We then extract the corresponding similarity scores from both the initial and updated similarity matrices using these fixed indices. This enables us to compare the distribution over a consistent local neighborhood before and after fine-tuning.

To preserve the local structural relationships, we impose a structure consistency constraint over the neighborhood similarity scores. Specifically, we define probability distributions over neighbors using a softmax function with temperature $\tau$:

\begin{equation}
    p(\mathcal{N}_p) = \text{softmax}\left( \mathbf{S}_p^{(\mathcal{N})} / \tau \right)
\end{equation}
where $p(\mathcal{N}_p)$ represents the probability distribution of the similarity scores over the selected Top-K neighbors. We then enforce consistency between the original and fine-tuned distributions using KL-divergence:

\begin{equation}
\begin{split}
\mathcal{L}_{\text{SCL}} &= 
\frac{1}{N_a} \sum_{i=1}^{N_a} D_{\text{KL}} \left( p\left(\mathbf{S}_a^{(0,\mathcal{N})}[i]\right) \parallel p\left(\mathbf{S}_a^{(+,\mathcal{N})}[i]\right) \right) \\
&\quad + 
\frac{1}{N_o} \sum_{i=1}^{N_o} D_{\text{KL}} \left( p\left(\mathbf{S}_o^{(0,\mathcal{N})}[i]\right) \parallel p\left(\mathbf{S}_o^{(+,\mathcal{N})}[i]\right) \right)
\end{split}
\end{equation}

where $N_a$ and $N_o$ are the number of seen attributes and objects. This loss encourages the learned embeddings to retain their original local structures, preventing fine-tuning from distorting the underlying semantic relationships.

\noindent\textbf{Structure-guided Adaptation Strategy.} In the Structure-aware Consistency Loss (SCL) process, we regularize the distributions of seen attributes and objects during training to preserve their original semantic relationships. However, while seen attributes and objects are progressively optimized to fit the task-specific data distribution, unseen ones remain fixed in the pretrained space. This discrepancy disrupts the local structure between seen and unseen concepts, making it challenging for unseen compositions to generalize effectively. To mitigate this issue, we introduce the Structure-guided Adaptation Strategy (SAS) at inference time, which dynamically refines the representations of unseen prompts by leveraging local structural information from seen attributes and objects.

Given a set of seen and unseen attributes and objects, we first obtain their initial text embeddings using the CLIP text encoder:

\begin{equation}
    \mathbf{t}_{p}^{(0)} = f_{\text{text}}(\mathbf{P}_{p}^{(0)}), \quad 
    \mathbf{t}_{p^*} = f_{\text{text}}(\mathbf{P}_{p^*}^{(0)})
\end{equation}

where $p \in \{a, o\}$ refers to either attribute or object, and $*$ denotes unseen attributes or objects. Next, we compute the similarity matrices between unseen and seen primitives:

\begin{equation}
    \mathbf{S}_p^{\text{u2s}} = \frac{\mathbf{t}_{p^*}^{(0)} \cdot (\mathbf{t}_{p}^{(0)})^\top}{\|\mathbf{t}_{p^*}^{(0)}\| \|\mathbf{t}_{p}^{(0)}\|}
\end{equation}

For each unseen primitive, we determine its Top-K most similar seen primitives:

\begin{equation} 
    \mathbf{S}_p^{\mathcal{N}} = \mathcal{N}_K(\mathbf{S}_p^{\text{u2s}})
\end{equation}

where $\mathcal{N}_K(\cdot)$ selects the Top-K highest similarity scores and their corresponding seen primitives. To update the representation of unseen primitives, we first compute the adjustment of seen primitives due to fine-tuning:

\begin{equation}
    \Delta \mathbf{P}_{p} = \mathbf{P}_{p}^{(+)} - \mathbf{P}_{p}^{(0)}
\end{equation}

Next, we obtain the normalized weights $w_k$ using a softmax function over the similarity scores:

\begin{equation}
    w_k = \frac{\exp(\mathbf{S}_p^{\mathcal{N}}[k] / \tau)}{\sum_{j \in \mathcal{N}_K} \exp(\mathbf{S}_p^{\mathcal{N}}[j] / \tau)}
\end{equation}

Using these weights, we compute the aggregated update for each unseen primitive:

\begin{equation}
    \Delta \mathbf{P}_{p^*} = \sum_{k \in \mathcal{N}_K} w_k \Delta \mathbf{P}_{p}[k]
\end{equation}

Finally, we update the unseen primitives accordingly:

\begin{equation}
    \mathbf{P}_{p^*}^{(+)} = \mathbf{P}_{p^*}^{(0)} + \Delta \mathbf{P}_{p^*}
\end{equation}

This adjustment ensures that unseen primitives adapt to the updated local structure of seen primitives while preserving their original semantics. 
{It is worth noting that while text-only embedding spaces may occasionally cluster visually distinct concepts (e.g., antonyms) due to contextual co-occurrence, CLIP’s multimodal pre-training promotes a strong alignment between semantic proximity and visual similarity. Quantitative evidence for this alignment is provided in \ref{4.5}}

\subsection{Training and Inference} \label{3.3}

\noindent\textbf{Training Objective.}
Our training objective consists of two components: (1) a classification loss to optimize compositional recognition, and (2) a Structure-aware Consistency Loss (SCL) that preserves local structure among seen primitives. The overall loss is defined as:
\begin{equation}
    \mathcal{L} = \mathcal{L}_{\text{CE}} + \lambda \mathcal{L}_{\text{SCL}},
\end{equation}
where $\lambda$ controls the trade-off between classification performance and structural regularization.

\noindent\textbf{Inference Procedure.}
At test time, we apply Structure-guided Adaptation Strategy (SAS) to adjust the prompt representations of unseen attributes and objects using structural information transferred from similar seen primitives. Formally, for each unseen primitive $p_u \in \{a_u, o_u\}$, we compute the calibrated prompt as:
\begin{equation}
    \mathbf{P}_{p^*}^{(+)} = \text{SAS}(\mathbf{P}_{p^*}^{(0)}),
\end{equation}
where $\text{SAS}(\cdot)$ denotes the Structure-guided Adaptation Strategy defined in Sec.~\ref{3.2}.

With the refined prompts for both seen and unseen primitives, we construct the full set of candidate composition prompts $\{\mathbf{P}_{y_i}\}_{y_i \in Y}$ and obtain their text embeddings $\{t_{y_i}\}_{y_i \in Y}$ using the CLIP text encoder.

Given a test image $x_i$, we compute its visual embedding $v_i = f_{\text{vis}}(x_i)$ and evaluate cosine similarity with each composition embedding $t_{y_j}$. The prediction probability is given by a temperature-scaled softmax:
\begin{equation}
    P(y_j | x_i) = \frac{\exp(S_{i,j} / \tau)}{\sum_{k=1}^{|Y|} \exp(S_{i,k} / \tau)},
\end{equation}
and the final prediction is:
\begin{equation}
    \hat{y} = \arg\max_{y_i \in Y} P(y_i \mid x_i).
\end{equation}

{\noindent\textbf{Algorithm Summary.}
{We present the complete training and inference procedure of our proposed SPA in Algorithm \ref{alg:spa}.}}

\begin{algorithm}[t]
\caption{Structure-aware Prompt Adaptation (SPA)}
\label{alg:spa}
\begin{algorithmic}[1]
\REQUIRE Seen primitive sets $A$, $O$; unseen primitive sets $A^*$, $O^*$; 
        CLIP encoders $f_{\text{vis}}$, $f_{\text{text}}$; hyperparameters $K$, $\tau$, $\lambda$.
\ENSURE Adapted prompts for all attribute-object compositions.

\STATE \textbf{Phase 1: Training with the Structure-aware Consistency loss}
\FOR{each primitive $p \in A \cup O$}
    \STATE Extract initial embedding $\mathbf{t}_p^{(0)}$ from CLIP
    \STATE Compute initial similarity matrix $\mathbf{S}_p^{(0)}$
    \STATE Identify Top-$K$ neighbor indices $\mathcal{I}_p$
\ENDFOR

\FOR{each training iteration}
    \FOR{each primitive $p \in A \cup O$}
        \STATE Extract optimized embedding $\mathbf{t}_p^{(+)}$
        \STATE Compute updated similarity matrix $\mathbf{S}_p^{(+)}$
        \STATE Extract neighborhood similarities $\mathbf{S}_p^{(0,\mathcal{N})}$, $\mathbf{S}_p^{(+,\mathcal{N})}$ using $\mathcal{I}_p$
    \ENDFOR
    \STATE Compute distributions over neighborhoods
    \STATE Calculate $\mathcal{L}_{\text{SCL}}$ using KL-divergence
    \STATE Minimizing $\mathcal{L} = \mathcal{L}_{\text{CE}} + \lambda\mathcal{L}_{\text{SCL}}$
\ENDFOR

\STATE \textbf{Phase 2: Inference with Structure-guided Adaptation Strategy}
\FOR{each primitive $p \in A \cup O$}
    \STATE Compute parameter shift $\Delta\mathbf{P}_p = \mathbf{P}_p^{(+)} - \mathbf{P}_p^{(0)}$
\ENDFOR

\FOR{each unseen primitive $p^* \in A^* \cup O^*$}
    \STATE Compute similarity between $p^*$ and seen primitives
    \STATE Retrieve Top-$K$ most similar seen neighbors $\mathcal{N}_K$
    \STATE Compute normalized weights $w_k$ using softmax
    \STATE Aggregate shift: $\Delta\mathbf{P}_{p^*} = \sum_{k \in \mathcal{N}_K} w_k \Delta\mathbf{P}_p[k]$
    \STATE Calibrate prompt: $\mathbf{P}_{p^*}^{(+)} = \mathbf{P}_{p^*}^{(0)} + \Delta\mathbf{P}_{p^*}$
\ENDFOR

\STATE Predict compositions using adapted prompts.
\end{algorithmic}
\end{algorithm}

\section{Experiments}
\subsection{Experiment Settings}
\begin{table*}[htbp]
\centering
\caption{\textbf{Comprehensive results on the MIT-States and C-GQA benchmarks.} The top block reports traditional CZSL methods, and the bottom block presents VLM-based prompt tuning approaches with and without SPA. Incorporating SPA consistently improves performance, notably yielding large gains on open-vocabulary splits, such as a +55.1\% relative improvement on $A^*O^*$ for C-GQA.}
\label{tab:mit&cgqa}
\scriptsize
\setlength{\tabcolsep}{2.5pt}
\begin{tabular}{l  *{9}{c}|*{9}{c}} 
\toprule
\multirow{2}{*}{Methods} & \multicolumn{9}{c|}{Mit-States} & \multicolumn{9}{c}{C-GQA} \\
\cmidrule(lr){2-10} \cmidrule(lr){11-19}
 &  HM$\uparrow$ & AUC$\uparrow$ & Seen$\uparrow$ & Unseen$\uparrow$ & $AO$$\uparrow$ & $(AO)^*$$\uparrow$ & $A^*O$$\uparrow$ & $AO^*$$\uparrow$ & $A^*O^*$$\uparrow$  
 &  HM$\uparrow$ & AUC$\uparrow$ & Seen$\uparrow$ & Unseen$\uparrow$ & $AO$$\uparrow$ & $(AO)^*$$\uparrow$ & $A^*O$$\uparrow$ & $AO^*$$\uparrow$ & $A^*O^*$$\uparrow$  
 \\
\midrule
\rowcolor{gray!19}
\multicolumn{19}{l}{\textit{Traditional Paradigm (Non-VLM)}} \\
LE~\cite{nagarajan2018attributes} & 7.64   & 1.01  & 16.29 & 9.46 & 10.24 & 11.38 & 5.98 & 4.15 & 2.87   & 8.39 & 1.17  & 19.37 & 8.36 & 10.76& 6.51 &9.53 & 2.67 & 1.08 \\
CompCos~\cite{mancini2021open} & 10.22  &1.97 & 26.53 & 10.29 & 14.32 & 21.09 & 5.86 & 2.89 & 0.63 & 9.64 & 2.35 & 40.19 & 7.25 & 20.19 & 20.24 & 4.47 & 1.95 & 0.26 \\
OADis~\cite{saini2022disentangling} & 9.55 & 1.83  & 25.35 & 10.79 & 12.18 & 16.06 & 6.40 & 5.41 & 1.34 & 9.74 & 2.33 & 42.88 & 7.12& 20.86 & 15.19 & 6.17 & 3.47 & 0.61 \\
SCEN~\cite{hao2023learning} & 9.72 & 1.73  & 22.08 & 8.25 & 11.85 & 30.02 & 3.82 & 0.33 &0.08 & 9.03& 1.97 &41.65 & 7.83 & 20.65 & 21.42 & 3.61 & 1.08 & 0.05 \\
CANet~\cite{wang2023learning} & 10.52 & 2.40  & 26.42 & 9.54 & 16.56 & 23.08 & 6.15 & 4.08 &0.58 &11.96  & 3.04 &40.52 & 9.21& 22.43 & 20.87 & 4.95 & 2.03 & 0.64 \\
OADis+NEL~\cite{saini2024beyond} & 10.94 & 2.41  & 29.02 & 11.13 & 14.11 & 18.87 & 8.24 & 5.49  & 3.54  & 12.11  & 3.18 & 42.38 & 9.77 & 19.78 & 16.07 & 12.86  & 2.87 & 3.04 \\
\cmidrule{1-19}
\rowcolor{gray!19}
\multicolumn{19}{l}{\textit{VLM Prompting Paradigm}} \\
CSP~\cite{DBLP:conf/iclr/NayakYB23}       & 26.82       & 10.22                     & 36.53           & 33.11              & 29.03          & 26.19             & 22.36           & 26.22          & 25.52   & 15.19     & 3.54                     & 29.47            & 14.03              & 19.54          & 10.87             & 16.49           & 7.65          & 9.28          \\
\rowcolor{blue!6}
\textbf{+SPA}  & \textbf{27.80}   & \textbf{10.59}                & 36.53            & \textbf{34.01}              & 28.47          & \textbf{27.90}             & \textbf{26.52}           & \textbf{27.38}          & \textbf{29.44}   & \textbf{16.70}           & \textbf{4.07}          & 29.47            & \textbf{15.77}              & \textbf{20.86}          & \textbf{13.11}         &\textbf{18.98}            &  7.12         & \textbf{10.67}           \\
\cmidrule{1-19}
HPL~\cite{wang2023hierarchical}              & 26.77           & 10.15          & 36.05            & 33.14              & 29.11          & 23.81             & 25.52           & 24.35          & 29.25    & 14.30           & 3.16         & 26.99            & 13.42              & 18.71          & 13.06             & 14.24           & 7.45          & 8.15          \\
\rowcolor{blue!6}
\textbf{+SPA}     & \textbf{27.49}           & \textbf{10.60}          & \textbf{36.45}            & \textbf{34.2}              & \textbf{29.6}          & \textbf{24.11}             & 25.48           & \textbf{25.71}          & \textbf{32.03}   & \textbf{16.21}           & \textbf{4.07}          & \textbf{32.12}            & \textbf{14.46}              & \textbf{22.19}          & \textbf{11.84}             & \textbf{17.47}           & 6.22          & \textbf{10.93}           \\
\cmidrule{1-19}
DFSP~\cite{DBLP:conf/cvpr/LuGLG23}      & 28.12        &12.09                     & 44.19            & 33.13              & 35.81          & 25.11             & 18.08           & 26.51          & 21.71     & 22.57     & 8.35             & 48.51           & 19.86      & 30.63            & 28.06              & 19.60          & 12.02           & 9.11          \\
\rowcolor{blue!6}
\textbf{+SPA}    & \textbf{28.90}  & \textbf{12.32}                    & 43.63            & \textbf{33.52}              & 34.11          & \textbf{27.36}             & \textbf{21.99}           & \textbf{26.88}          &\textbf{24.07}   & \textbf{22.81}           & \textbf{8.51}          & \textbf{48.68}            & \textbf{20.24}              & 30.63          & \textbf{29.03}             & 19.50           &  \textbf{12.21}         & \textbf{10.75}           \\

\cmidrule{1-19}
Troika~\cite{huang2024troika}        & 27.11           & 10.89                & 39.60            & 32.70              & 30.32          & 28.45             & 21.78           & 26.41          & 18.16    & 20.52           & 6.90    & 43.87          & 18.08       & 27.98            & 22.91              & 19.40          & 11.74             & 1.73                    \\
\rowcolor{blue!6}
\textbf{+SPA}     & \textbf{27.45}           & \textbf{10.97}          & 39.03            & \textbf{33.56}              & 30.32          & 26.16             & \textbf{24.15}           & 26.22          & \textbf{26.15}   & \textbf{21.43}           & \textbf{7.03}          & 40.89            & \textbf{19.49}              & 27.81          & 18.45             & \textbf{21.91}           & 11.74          & \textbf{11.54}           \\
\cmidrule{1-19}
Baseline Avg    & 27.21 & 10.84 & 39.09 & 33.02 & 31.07 & 25.89 & 21.94 & 25.87 & 23.66 & 18.15 & 5.49 & 37.21 & 16.35 & 24.22 & 18.73 & 17.43 & 9.72 & 7.07 \\
\rowcolor{blue!6}
\textbf{+SPA Avg} & \textbf{27.91} & \textbf{11.12} & 38.91 & \textbf{33.82} & 30.63 & \textbf{26.38} & \textbf{24.54} & \textbf{26.55} & \textbf{27.92} & \textbf{19.29} & \textbf{5.92} & \textbf{37.79} & \textbf{17.49} & \textbf{25.37} & 18.11 & \textbf{19.47} & 9.32 & \textbf{10.97} \\
\bottomrule
\end{tabular}

\end{table*}

\noindent{\textbf{Datasets.}}
To thoroughly assess the generalization and robustness of our method, we conduct experiments on 
{four OV-CZSL benchmarks, \ie MIT-States~\cite{isola2015discovering}, C-GQA~\cite{mancini2022learning}, VAW-CZSL~\cite{pham2021learning} and UT-Zappos~\cite{yu2014fine}}.
The MIT-States dataset includes 115 attributes and 245 objects, yielding 1,962 attribute–object compositions and a total of 53,753 images. C-GQA, derived from Stanford GQA, expands this scale with 413 attributes and 674 objects, forming roughly 7,000 pairs with 39,298 samples. The largest benchmark, VAW-CZSL, consists of 533 attributes and 543 objects, leading to 15,785 combinations and 92,000 images in total. 
{UT-Zappos comprises 50,025 fine-grained shoe images, encompassing 116 unique compositions derived from 16 attributes and 12 objects.}
We follow the dataset split proposed by ~\citet{saini2024beyond} to ensure a fair evaluation under the open-vocabulary setting. 
{As no established open-vocabulary partition exists for UT-Zappos, we define a new split for this benchmark following the same rigorous criteria to ensure consistency across all evaluations.}
Detailed statistics of these partitions, including the numbers of seen/unseen attributes and objects and the  composition types ($AO$, $(AO)^*$, $A^*O$, \etc), are provided in the supplementary material.

\noindent{\textbf{Evaluation Metrics.}}
Following established protocols~\cite{li2020symmetry,saini2022disentangling,mancini2021open,purushwalkam2019task}, our evaluation provides a multi-faceted assessment of model performance.
To measure the overall balance between seen ($Y^s$) and unseen ($Y^u$) compositions, we report the Harmonic Mean (HM). Additionally, we use the Area Under the Curve (AUC) to evaluate the model's robustness to different inference biases. We primarily report Top-1 AUC, and Top-3 AUC is used for the large and challenging VAW-CZSL benchmark. 
For a more granular analysis of generalization, we further provide a detailed breakdown of accuracies on five key subsets: fully seen compositions ($AO$), unseen combinations of seen primitives($(AO)^*$), and the three challenging open-vocabulary cases ($A^*O$, $AO^*$, and $A^*O^*$).

\noindent{\textbf{Baseline.}} To evaluate the effectiveness of our SPA, we integrate it with a set of representative CZSL baselines. These include single-branch methods, such as CSP~\cite{DBLP:conf/iclr/NayakYB23} and DFSP~\cite{DBLP:conf/cvpr/LuGLG23}, as well as multi-branch architectures, including HPL~\cite{wang2023hierarchical} and Troika~\cite{huang2024troika}.

\noindent{\textbf{Implementation Details.}} 
Following prior works~\cite{DBLP:conf/iclr/NayakYB23,wang2023hierarchical,DBLP:conf/cvpr/LuGLG23,huang2024troika}, we use CLIP as the vision–language backbone. To reduce computational overhead, we adopt CLIP ViT-B/32 across all experiments instead of the commonly used CLIP-L/14. Importantly, all models are trained under the same backbone and protocol, ensuring a fair comparison. 
Both the image and text encoders are initialized from the pre-trained CLIP ViT-B/32 model and keep frozen during training. For optimization, we use AdamW with a learning rate of $5 \times 10^{-5}$ for MIT-States and C-GQA, and $1 \times 10^{-5}$ for VAW-CZSL, together with a weight decay of $1 \times 10^{-6}$. All models are trained for 20 epochs. 
Regarding training hyperparameters, the batch size is set to 256 for MIT-States and 128 for C-GQA and VAW-CZSL. We fix the number of neighbors $K$ at 5 and the structure consistency weight $\lambda$ at 1.
{Regarding the Structure-guided Adaptation Strategy (SAS), the Top-K neighbor search is performed independently for each dataset, restricted solely to the seen primitives within that dataset’s respective training split. Furthermore, as specified in our formulations, all attribute and object embeddings are $L_2$-normalized prior to calculating cosine similarities during both the training and inference stages. }

\begin{table}[htbp]
\centering
\scriptsize
\caption{\textbf{Results on the VAW-CZSL benchmark.} Compared with traditional CZSL methods, VLM-based prompting approaches achieve stronger generalization. 
Incorporating SPA further enhances performance, particularly under fine-grained and open-vocabulary settings, showing consistent gains across most metrics.}
\resizebox{\linewidth}{!}{
\begin{tabularx}{\linewidth}{@{}l *{8}{>{\centering\arraybackslash}X}@{}}
\toprule
\multirow{2}{*}{Methods} & \multicolumn{7}{c}{VAW-CZSL} \\
\cmidrule(lr){2-8}
& HM$\uparrow$ & AUC$\uparrow$ & $AO$$\uparrow$ & $(AO)^*$$\uparrow$ & $A^*O$$\uparrow$ & $AO^*$$\uparrow$ & $A^*O^*$$\uparrow$ \\
\midrule
\rowcolor{gray!19}
\multicolumn{8}{l}{\textit{Traditional Paradigm (Non-VLM)}} \\
LE~\cite{nagarajan2018attributes} & 8.27   & 1.49  & 15.62 & 10.48 & 5.79 & 2.78 & 0.98  \\
CompCos~\cite{mancini2021open} & 10.68  &2.69 & 20.21 & 20.58 & 5.04 & 2.48 & 0.5  \\
OADis~\cite{saini2022disentangling} & 10.91 & 2.68  & 21.19 & 15.65 & 6.75 & 3.16 & 0.76   \\
SCEN~\cite{hao2023learning} & 10.64 & 2.53  & 19.06 & 20.76 & 4.52 & 2.05 &0.42 \\
CANet~\cite{wang2023learning} & 11.21 & 2.89  & 24.56 & 18.42 & 5.74 & 2.86 & 0.95  \\
OADis+NEL~\cite{saini2024beyond} & 11.35 & 2.91  & 23.02 & 16.18 & 7.86 & 3.37 & 1.36 \\
\midrule
\rowcolor{gray!19}
\multicolumn{8}{l}{\textit{VLM Prompting Paradigm}} \\
CSP~\cite{DBLP:conf/iclr/NayakYB23}           & 13.26 & 2.71 & 12.95 & 4.43 & 3.21 & 5.55 & 5.59\\
\rowcolor{blue!6}
\textbf{+SPA}       & \textbf{13.92} & 2.42 & 9.75 & \textbf{6.91} & \textbf{4.36} & \textbf{6.40} & \textbf{6.70}\\
\cmidrule{1-8}
HPL~\cite{wang2023hierarchical}              & 11.60 & 2.12 & 9.44 & 4.18 & 2.42 & 4.36 & 4.18\\
\rowcolor{blue!6}
\textbf{+SPA}      & \textbf{13.03} & \textbf{2.45} & \textbf{11.38} & \textbf{4.94} & \textbf{2.74} & \textbf{4.81} & \textbf{5.07}\\
\cmidrule{1-8}
DFSP~\cite{DBLP:conf/cvpr/LuGLG23}              & 20.87 & 7.64 & 26.57 & 7.26 & 3.24 & 6.05  & 4.78\\
\rowcolor{blue!6}
\textbf{+SPA}       & \textbf{21.91} & \textbf{8.14} & 26.39 & \textbf{8.38} & \textbf{3.61} & \textbf{6.87} & \textbf{5.29} \\
\cmidrule{1-8}
Troika~\cite{huang2024troika}  & 18.25 & 5.17 & 24.94 & 7.42& 2.09 & 6.71  & 0.72\\
\rowcolor{blue!6}
\textbf{+SPA}       & \textbf{20.34} & \textbf{5.26} & 21.79 & \textbf{9.16} & \textbf{5.07} & \textbf{7.32} & \textbf{3.24}\\
\cmidrule{1-8}
Baseline Avg    & 16.00 & 4.41 & 18.48 & 5.82 & 2.74 & 5.67 & 3.82 \\
\rowcolor{blue!6}
\textbf{+SPA Avg}  & \textbf{17.30} & \textbf{4.56} & 17.33 & \textbf{7.35} & \textbf{3.95} & \textbf{6.35} &\textbf{5.08}\\
\bottomrule
\label{tab:vaw}

\end{tabularx}
}
\vspace{-1.5em}
\end{table}

\subsection{Main Results}

\noindent{\textbf{Results on MIT-States and C-GQA.}}
We first evaluate our method on two OV-CZSL benchmarks, MIT-States and C-GQA, and report results in Tab.~\ref{tab:mit&cgqa}. 
Methods are grouped into two paradigms: (1) traditional CZSL approaches without vision–language pretraining (top block), and (2) vision–language model (VLM)-based prompt tuning methods (bottom block). For the latter, we assess performance both with and without our SPA.

As shown in Tab.~\ref{tab:mit&cgqa}, traditional non-VLM methods (\eg LE~\cite{nagarajan2018attributes}, CompCos~\cite{mancini2021open}, OADis~\cite{saini2022disentangling}) generally underperform on both overall metrics (HM, AUC) and composition-level metrics. In contrast, VLM-based prompt tuning approaches obtain substantially better baselines due to rich pre-trained representations and flexible prompt designs, though they still struggle on compositions containing unseen primitives.
Within this VLM paradigm, integrating SPA consistently yields improvements across four strong baselines. 
On the MIT-States dataset, we first validate this effectiveness. As shown in Tab.~\ref{tab:mit&cgqa}, SPA achieves consistent improvements across the board, increasing the overall HM and AUC scores by +2.6\%. More importantly, it shows strong generalization to open-vocabulary compositions, raising the $A^*O$ and $A^*O^*$ metrics by +11.9\% and +18.0\%, respectively.
This generalization ability is even more pronounced on the more challenging C-GQA dataset. SPA boosts the overall HM by +6.3\% and the AUC by +7.8\%. Its most notable impact is on the hardest split, $A^*O^*$, where it achieves a +55.1\% relative gain (from 7.07 to 10.97), highlighting its capability to handle entirely unseen compositions.
{Note that while SPA delivers substantial gains in targeted unseen splits, the improvement in AUC is more measured. This stems from the nature of AUC as a global integral across the entire bias spectrum. Consequently, within the aggregate AUC calculation, the marginal reduction in seen performance partially offsets the sharp gains on the unseen splits.}
We also note a slight drop on $AO^*$ for C-GQA (9.72 $\to$ 9.32), which likely stems from some unseen objects being semantically/visually distant from the seen set; we analyze these failure modes and provide qualitative examples in Sec.~\ref{sec_vis}. 
Overall, SPA preserves competitive performance on seen compositions while substantially improving generalization to unseen ones, demonstrating its effectiveness for open-vocabulary CZSL.


\noindent{\textbf{Results on VAW-CZSL.}}
To further validate the effectiveness of our approach, we conduct additional evaluations on the large-scale and highly challenging VAW-CZSL dataset, with results presented in Tab.~\ref{tab:vaw}. Following prior work~\cite{saini2024beyond}, all metrics are reported under the Top-3 AUC setting due to the dataset’s difficulty. 
The results also confirm the VLM paradigm's superiority on this complex dataset, where our SPA-enhanced models set a new state-of-the-art. 
By consistently improving performance across four representative baselines, SPA raises the average HM from 16.00 to 17.30 and the AUC from 4.41 to 4.56. Beyond overall gains, SPA delivers notable improvements on open-vocabulary splits, including $A^*O$ (+44.1\%), $AO^*$ (+12.0\%), and $A^*O^*$ (+33.0\%). 
{While a marginal absolute decrease is observed in $AO$ performance (18.48 $\rightarrow$ 17.33), we interpret this as a strategic trade-off to mitigate the co-occurrence bias inherent in seen pairs. Specifically, by slightly relaxing the empirical fitting on the training distribution, SPA achieves substantial gains across all open-vocabulary splits. The consistent improvement in Harmonic Mean (HM) confirms that this structural regularization provides a net benefit for global model robustness.} Overall, SPA provides a plug-and-play enhancement for VLM-based CZSL methods, maintaining strong performance in open-vocabulary settings even on challenging, large-scale datasets.

\noindent{\textbf{{Results on UT-Zappos.}}}
{Beyond large-scale benchmarks, we evaluate SPA on the fine-grained shoe images of UT-Zappos. Given the absence of prior open-vocabulary settings for this dataset, we defined a new, rigorous partition for a comprehensive assessment. The results in Table~\ref{tab:utzappos_results} consistently demonstrate that SPA enhances performance across a variety of metrics. Specifically, for the baseline average, SPA achieves a +2.02 absolute gain in HM and a substantial +7.72 gain (over 4$\times$ improvement) in $A^*O^*$. These results highlight our method's robustness when dealing with subtle, fine-grained attribute differences in novel compositions.}

\begin{table}[tbp]
\centering
\scriptsize
\caption{{\textbf{Comprehensive results on the UT-Zappos benchmarks.} 
Integrating SPA consistently bolsters performance across various baselines, particularly in open-vocabulary scenarios. Such empirical evidence underscores the efficacy of our structural reasoning in capturing nuanced attribute-object relationships.}}
\begin{tabularx}{\linewidth}{@{}l *{7}{>{\centering\arraybackslash}X}@{}}
\toprule
\multirow{2}{*}{Methods} & \multicolumn{7}{c}{UT-Zappos} \\
\cmidrule(lr){2-8}
& HM$\uparrow$ & AUC$\uparrow$ & $AO$$\uparrow$ & $(AO)^*$$\uparrow$ & $A^*O$$\uparrow$ & $AO^*$$\uparrow$ & $A^*O^*$$\uparrow$ \\
\midrule
\rowcolor{gray!19}
\multicolumn{8}{l}{\textit{VLM Prompting Paradigm}} \\
CSP~\cite{DBLP:conf/iclr/NayakYB23}            & 27.33 & 9.56 & \textbf{16.01} & 8.58 & 3.97 & 7.16 & 3.33    \\
\rowcolor{blue!6}
\textbf{+SPA}  & \textbf{29.01} & \textbf{10.75} & 15.91 & \textbf{8.86} & \textbf{5.66} & \textbf{9.66} & \textbf{10.57} \\
\cmidrule{1-8}
HPL~\cite{wang2023hierarchical}            & 27.83 & 9.70 & \textbf{16.52} & \textbf{9.10} & 3.23 & 6.56 & 1.98 \\
\rowcolor{blue!6}
\textbf{+SPA}  & \textbf{30.13} & \textbf{11.03} & 15.99 & 8.99 & \textbf{6.32} & \textbf{7.38} & \textbf{8.67} \\
\cmidrule{1-8}
DFSP~\cite{DBLP:conf/cvpr/LuGLG23}           & 32.07 & 15.00 & \textbf{20.86} & 13.77 & 6.61 & 10.39  & 2.38 \\
\rowcolor{blue!6}
\textbf{+SPA}  & \textbf{34.32} & \textbf{16.83} & 20.64 & \textbf{14.40} & \textbf{7.18} & \textbf{11.64} & \textbf{13.69} \\
\cmidrule{1-8}
Troika~\cite{huang2024troika}         & 31.13 & 14.88 & \textbf{21.32} & 10.87 & 5.47 & 7.83 & 2.11 \\
\rowcolor{blue!6}
\textbf{+SPA}  & \textbf{32.98} & \textbf{15.42} & 20.93 & \textbf{11.03} & \textbf{6.85} & \textbf{9.11} & \textbf{7.73} \\
\cmidrule{1-8}
Baseline Avg   & 29.59 & 12.29 & \textbf{18.68} & 10.58 & 4.82 & 7.98 & 2.45 \\
\rowcolor{blue!6}
\textbf{+SPA Avg} & \textbf{31.61} & \textbf{13.51} & 18.37 & \textbf{10.82} & \textbf{6.51} & \textbf{9.45} & \textbf{10.17} \\
\bottomrule
\label{tab:utzappos_results}
\end{tabularx}
\vspace{-1.5em}
\end{table}

\subsection{Ablation Study}

In this section, we conduct detailed ablation studies to investigate the contribution of each component. All experiments are conducted on the MIT-States~\cite{isola2015discovering} dataset, using CSP~\cite{DBLP:conf/iclr/NayakYB23} as the baseline model. We primarily report overall performance metrics (HM and AUC), and then place particular emphasis on open-vocabulary metrics ($(A^*O, AO^*, A^*O^*$).

\noindent{\textbf{Effectiveness of components.}}
To validate the individual effectiveness of our two core components, the Structure-aware Consistency Loss (SCL) and the Structure-guided Adaptation Strategy (SAS), we conduct an incremental ablation study with results presented in Tab.~\ref{tab:ablation_components_full}.
Our baseline model (CSP) achieves an HM of 26.82 and an $A^*O^*$ score of 25.52. When individually integrating our proposed components, both yield clear benefits. Specifically, incorporating SAS alone increases HM from 26.82 to 27.52 and the challenging $A^*O^*$ score from 25.52 to 28.05. SCL alone also provides a clear benefit, improving the HM to 27.14. Importantly, combining both components in our full SPA model yields the best overall performance. The full model achieves the highest scores across all key metrics, reaching an HM of 27.80 and $A^*O^*$  score of 29.44. This analysis demonstrates that SCL and SAS are complementary, and both are essential for achieving the optimal performance of SPA.

\begin{table}[t]
\centering
\small
\caption{\textbf{Ablation study on the effect of the number of neighbors ($K$) in SPA.} We report key performance metrics on MIT-States, showing that a moderate choice of $K=5$ achieves the best trade-off across HM and open-vocabulary metrics.}
\label{tab:ablation_k}
\begin{tabular}{@{} c | S[table-format=2.2] S[table-format=2.2] S[table-format=2.2] S[table-format=2.2] S[table-format=2.2] @{}}
\toprule
\multirow{2}{*}{\textbf{Neighbors $K$}} & \multicolumn{5}{c}{MIT-States} \\
& {HM$\uparrow$} & {AUC$\uparrow$} & {\textbf{$A^*O$$\uparrow$}} & {\textbf{$AO^*$$\uparrow$}} & {\textbf{$A^*O^*$$\uparrow$}} \\
\midrule
$K = 0$ (Baseline) & 26.82 & 10.22 & 22.36 & 26.22 & 25.52 \\
$K = 1$ & 27.59 & 10.64 & 27.41 & 26.24 & 29.44 \\
$K = 3$ & 27.70 & 10.66 & 26.43 & 27.30 & 29.25 \\
\rowcolor{blue!6}
$K = 5$ & \textbf{27.80} & 10.59 & \textbf{26.52} & \textbf{27.38} & \textbf{29.44} \\
$K = 7$ & 27.70 & \textbf{10.66} & 26.43 & 27.30 & 29.25 \\
$K = 10$ & 27.78 & 10.66 & 26.41 & 27.35 & 29.31 \\
\bottomrule
\end{tabular}
\vspace{1em}
\end{table}

\noindent{\textbf{Effect of number of Top-$K$ Neighbors.}}
We investigate the impact of the nearest neighbor count (K), which controls the amount of local structural information used in both SCL and SAS modules. Tab.~\ref{tab:ablation_k} shows that our method is robust across a wide range of $K$ values, consistently outperforming the baseline. Here, $K = 0$ corresponds to the baseline, where no local structure is leveraged for adaptation. The best overall trade-off is achieved at K=5, which yields the highest HM (27.80) and $A^*O^*$ (29.44). We observe that a small $K$ (\eg 1 or 3) already provides meaningful structure propagation, while a very large K (\eg 10) may introduce less relevant neighbors that slightly dilute the semantic signal. This confirms that $K = 5$ provides an optimal balance between capturing a rich local structure and avoiding potential noise. These results indicate that our approach effectively exploits local structural information while maintaining robustness to the choice of $K$.

{While $K = 5$ is initially determined on MIT-States, we further evaluate its scalability across larger datasets including C-GQA and VAW-CZSL. As detailed in the Supplementary Material, SPA exhibits remarkable stability, as $K=5$ consistently emerges as the optimal sweet spot despite varying vocabulary densities. This confirms that our structure-aware adaptation is primarily governed by intrinsic semantic locality, which implies that a concept's meaningful neighbors are naturally limited and independent of global vocabulary statistics.}

\noindent{\textbf{Effect of hyper-parameters $\lambda$ and $\tau$.}} 
We further analyze the sensitivity of the loss weight of the SCL $\lambda$ and the temperature $\tau$. 
{Values that are too high over-constrain the structure, while values too low allow the fine-tuning to distort the useful prior knowledge. The weight $\lambda$ balances the primary classification objective with our structural consistency regularizer. The temperature $\tau$ controls the sharpness of the neighbor similarity distribution. Lower temperatures (\eg 0.01) can make the loss brittle, while higher temperatures (\eg 0.20) over-smooth the distribution and include less relevant concepts.}
Our SPA demonstrates robustness across a wide range of values for both hyper-parameters. Detailed analyzes and results are provided in the supplementary material, showing optimal performance at $\lambda = 1.00$ and $\tau = 0.10$.

\newcommand{\gains}[1]{\textcolor{blue}{(+#1)}}

\begin{table*}[htbp]
\centering
\caption{\textbf{Component-wise computational cost and performance trade-off analysis of SPA on the MIT-States dataset.} The values in parentheses denote the increase in cost or performance over the CSP baseline.}
\label{tab:computational_cost}

\begin{tabular*}{0.85\textwidth}{@{\extracolsep{\fill}} l | cc | cc | cc @{}}
\toprule
& \multicolumn{2}{c|}{Training Overhead} & \multicolumn{2}{c|}{Inference Overhead} & \multicolumn{2}{c}{Performance Gain} \\
Method & {\shortstack{Time (min) $\downarrow$}} & {\shortstack{Memory (MB) $\downarrow$}} & {\shortstack{Speed (ms/img) $\downarrow$}} & {\shortstack{Memory (MB) $\downarrow$}} & {\shortstack{HM$\uparrow$}} & {\shortstack{A*O*$\uparrow$}} \\
\midrule
Baseline (CSP) & 20.33 & 6388 & 2.34 & 4560 & 26.82 & 25.52 \\
\midrule
\quad \textit{w} SCL & 21.38 \gains{1.05} & 6490 \gains{102} & 2.34 \gains{0.00} & 4560 \gains{0.00} & 27.14 \gains{0.32} & 26.15 \gains{0.63} \\
\quad \textit{w} SAS & 20.33 \gains{0.00} & 6388 \gains{0.00} & 2.36 \gains{0.02} & 4574 \gains{0.14} & 27.52 \gains{0.70} & 28.05 \gains{2.53} \\
\midrule

\textbf{SPA (Full)} & \textbf{21.38 \gains{1.05}} & \textbf{6490 \gains{102}} & \textbf{2.36 \gains{0.02}} & \textbf{4574 \gains{14}} & \textbf{27.80 \gains{0.98}} & \textbf{29.44 \gains{3.92}} \\
\bottomrule
\end{tabular*}
\end{table*}


\subsection{Computational Cost Analysis}
In this section, we analyze the computational overhead of our proposed Structure-aware Prompt Adaptation (SPA) method to demonstrate its efficiency and practicality. We evaluate its impact on training and inference. Specifically, We dissect the cost of each component in SPA when applied to the CSP baseline, with results detailed in Tab.~\ref{tab:computational_cost}. The Structure-aware Consistency Loss (SCL) is applied during training, introducing a minimal overhead. Specifically, it increases the training time by only 1.05 minutes ($\sim$5.2\%) and training memory by 102MB ($\sim$1.6\%). The Structure-guided Adaptation Strategy (SAS) operates during inference, adding a negligible 0.02ms to the per-image inference time and 14MB of memory. When combined, our full SPA module enhances performance significantly while maintaining high efficiency, proving its excellent cost-benefit trade-off.

We also considered the module's scalability for extremely large vocabularies. A detailed analysis is provided in the supplementary material, which identifies the k-NN search operation within our SAS module as the primary bottleneck ($>80\%$ of overhead). We note that k-NN search can be readily replaced by more efficient Approximate Nearest Neighbor (ANN) algorithms (\eg Faiss, ScaNN) to enhance SPA's applicability to massive-scale vocabularies.

\begin{figure*}[ht]
    \centering
    \includegraphics[width=\linewidth]{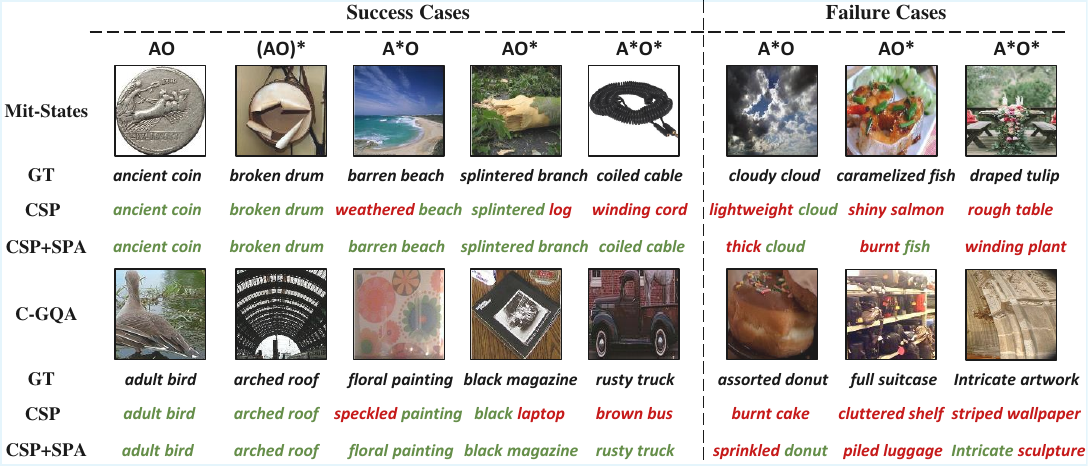}
    \caption{\textbf{Qualitative comparison of CSP+SPA against the baseline CSP on samples from MIT-States (top) and C-GQA (bottom).} GT denotes Ground Truth. SPA corrects the baseline's (CSP) prediction errors on challenging open-vocabulary compositions ($A^*O$, $AO^*$, $A^*O^*$) while maintaining strong performance on seen ones ($AO$, $(AO)^*$). Failure cases for open-vocabulary types are also shown to analyze the method's boundaries. Predictions marked in \textbf{\textcolor{MyDarkRed}{red}} are incorrect; those in \textbf{\textcolor{MyDarkGreen}{green}} are correct.}
    \label{vis1}
\end{figure*}

\subsection{In-depth Discussion and Analysis}  \label{4.5}
\noindent\textbf{Analysis of Design Choices.}
To justify our design choices, we conduct additional experiments on two core aspects: (1) adopting prompt tuning instead of full fine-tuning, and (2) evaluating our proposed SPA mechanism against the prior Neighborhood Expansion Loss (NEL)~\cite{saini2024beyond}. Since NEL was originally developed for traditional CZSL methods, we implement it on the same VLM baseline (CSP+NEL) for a fair comparison.

\begin{table}[htbp]
\centering
\small
\caption{\textbf{Comparison of design choices on MIT-States.} CSP + SPA outperforms full fine-tuning CLIP, CSP, and CSP + NEL across all metrics while maintaining low training memory.}
\label{tab:design_choice_comparison}
\resizebox{\columnwidth}{!}{%
\begin{tabular}{@{} l S[table-format=2.2] S[table-format=2.2] S[table-format=2.2] S[table-format=2.2] S[table-format=5.0] @{}}
\toprule
\multirow{2}{*}{Method} & \multicolumn{5}{c}{MIT-States} \\
& {HM$\uparrow$} & {$A^*O$$\uparrow$} & {$AO^*$$\uparrow$} & {$A^*O^*$$\uparrow$} & {Train Memory$\downarrow$}\\
\midrule
CLIP (Full FT) & 25.09 & 22.36 & 25.73 & 24.95 & 22748 \\
CSP & 26.82 & 22.36 & 26.22 & 25.52 & \textbf{6388} \\
CSP + NEL & 26.93 & 24.43 & 25.52 & 26.47 & 11544 \\
\rowcolor{blue!6}
CSP + SPA & \textbf{27.80} & \textbf{26.52} & \textbf{27.38} & \textbf{29.44} & 6490 \\
\bottomrule
\end{tabular}%
}
\end{table}

\subsubsection*{1) Comparison with Full Fine-tuning}
We first examine the choice of using a frozen backbone with prompt tuning. Tab.~\ref{tab:design_choice_comparison} shows that full fine-tuning of CLIP (denoted as CLIP (Full FT)) leads to a degraded performance (\eg HM from 26.82 for CSP down to 25.09) and incurs a nearly 4× increase in training memory. In contrast, frozen-backbone prompt tuning (denoted as CSP) preserves pre-trained representations and adapts efficiently to OV-CZSL, achieving a better trade-off between accuracy and efficiency.

\subsubsection*{2) Comparison with NEL}
Next, we compare SPA with the prior Neighborhood Expansion Loss. CSP+SPA consistently outperforms CSP+NEL across all key metrics on MIT-States. For instance, it improves the HM from 26.93 to 27.80 and achieves a substantial gain on the challenging $A^*O^*$ split (29.44 \vs 26.47). Moreover, SPA attains these improvements while using roughly half the training memory of CSP+NEL. This demonstrates that SPA provides a more effective and efficient mechanism for structured adaptation within the VLM prompting paradigm.

\begin{table*}[htbp]
\centering
\small
\caption{\textbf{Ablation study of SPA components on MIT-States~\cite{isola2015discovering} using the CSP~\cite{DBLP:conf/iclr/NayakYB23} baseline.} We incrementally add the Structure-aware Consistency Loss (SCL) and Structure-guided Adaptation Strategy (SAS), showing that each component improves performance, with the full SPA model achieving the best overall results across HM, AUC, and open-vocabulary metrics.}
\label{tab:ablation_components_full}
\begin{tabularx}{0.8\textwidth}{@{} l | cc | *{5}{>{\centering\arraybackslash}X}  *{5}{>{\centering\arraybackslash}X} @{}}
\toprule
\multirow{2}{*}{Setting} & \multicolumn{2}{c|}{Components} & \multicolumn{5}{c}{MIT-States}  \\
& SCL & SAS & HM$\uparrow$ & AUC$\uparrow$ & $A^*O$$\uparrow$ & $AO^*$$\uparrow$ & $A^*O^*$$\uparrow$ \\
\midrule
Baseline (\textit{w/o} SPA) & & & 26.82 & 10.22 & 22.36 & 26.22 & 25.52  \\
+SCL & \checkmark & & 27.14 & 10.34 & 23.73 & 27.33 & 26.15  \\
+SAS & & \checkmark & 27.52 & 10.58 & 22.94 & 26.83 & 28.05  \\
\rowcolor{blue!6} 
\textbf{+SCL +SAS (Full SPA)} & \checkmark & \checkmark & \textbf{27.80} & \textbf{10.59} & \textbf{26.52} & \textbf{27.38} & \textbf{29.44} \\
\bottomrule
\end{tabularx}
\vspace{-1em}
\end{table*}

\noindent{\textbf{Validity of Semantic-Visual Alignment.} 
To investigate the reliability of using semantic proximity as a proxy for visual similarity, we conduct a quantitative alignment analysis on the MIT-States dataset. Specifically, after applying feature mean-centering to account for CLIP’s inherent anisotropy, we evaluate the consistency between the semantic and visual spaces. As shown in Table~\ref{tab:correlation_compact} and Table~\ref{tab:consistency}, the highly significant Pearson correlation ($p = 0.0026 < 0.01$) and the 13.7$\times$ higher visual consistency of Top-5 semantic neighbors compared to random baselines demonstrate that semantic proximity in CLIP is a robust, non-random proxy for visual similarity. This evidence suggests that CLIP's multimodal pre-training effectively mitigates the risk of semantic-visual misalignment (\eg antonym interference), providing high-fidelity structural signals for SAS.}

\begin{table}[h]
\centering
\caption{{Quantitative Analysis of Semantic-Visual Alignment.}}
\label{tab:correlation_compact}
\begin{tabular}{lccc}
\toprule
\textbf{Correlation Measure} & \textbf{Pearson ($\rho$)} & \textbf{Spearman ($r_s$)} & \textbf{P-value} \\ \midrule
Semantic vs. Visual Distance & \textbf{0.1548} & \textbf{0.2153} & \textbf{0.0026} \\ \bottomrule
\end{tabular}
\vspace{-1em}
\end{table}

\begin{table}[h]
\centering
\caption{{Consistency Check: Visual Similarity of Semantic Neighbors}}
\label{tab:consistency}
\begin{tabular}{lc}
\toprule
\textbf{Neighbor Selection} & \textbf{Avg. Visual Similarity (Mean-Centered)} \\ \midrule
Random Neighbors            & 0.0047                                         \\
\textbf{Top-5 Semantic Neighbors} & \textbf{0.0642 ($\uparrow$ 13.7$\times$)} \\ \bottomrule
\end{tabular}
\end{table}

\begin{figure}[t]
  \centering
  \includegraphics[width=0.85\linewidth]{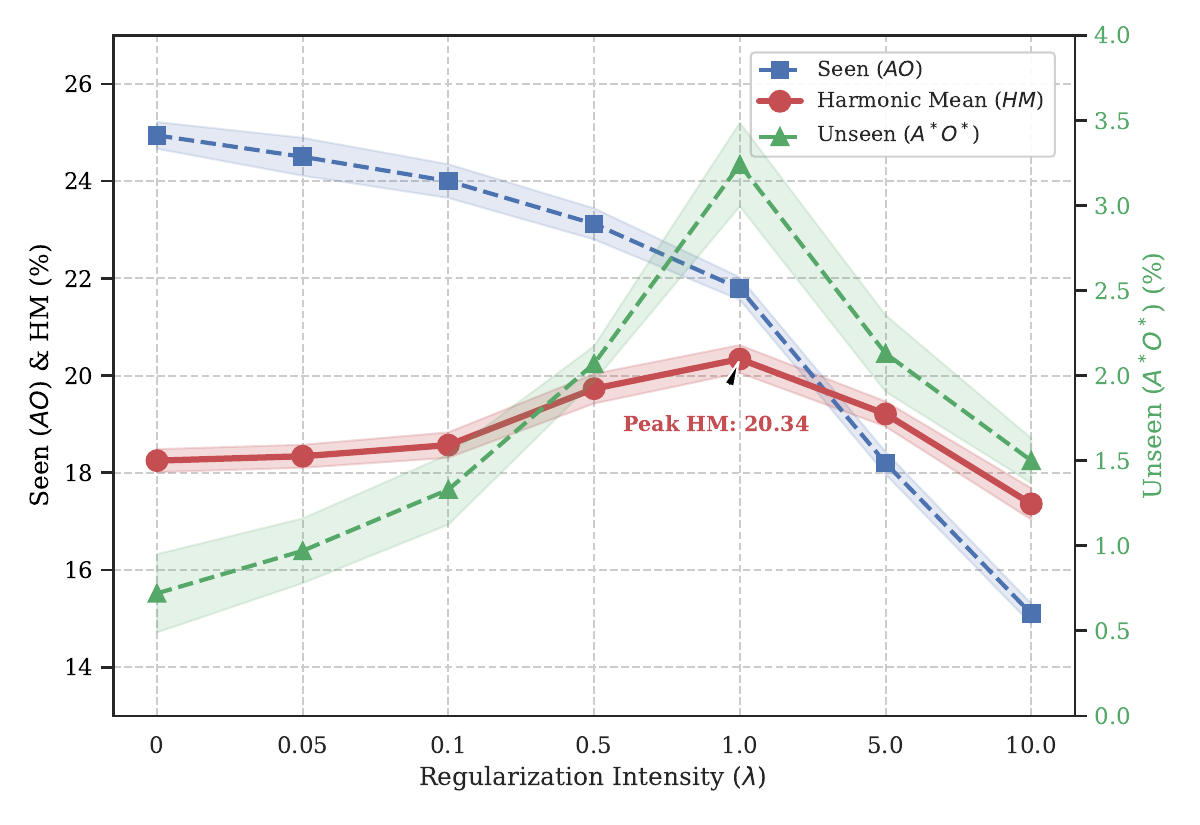} 
    \caption{{\textbf{Sensitivity analysis of the regularization intensity $\lambda$ for Troika on the VAW-CZSL dataset.}
  To ensure transparency and robustness, we report the average performance across \textbf{five independent runs with different random seeds}. 
  As $\lambda$ increases, the marginal decrease in Seen performance ($AO$) is significantly outweighed by the sharp gain in Unseen performance ($A^*O^*$), resulting in the \textbf{Harmonic Mean ($HM$) reaching its peak at $\lambda=1.0$}. 
  This confirms that $\lambda=1.0$ is the optimal equilibrium point for balancing empirical fitting and structural manifold preservation.}}
  \label{fig:lambda_sensitivity}
\end{figure}

\noindent{\textbf{Impact of Structural Regularization Intensity.} }
{To investigate whether our structural regularization is overly restrictive, we conduct a sensitivity analysis of the intensity parameter $\lambda$ using Troika+SPA on the VAW-CZSL dataset. As shown in Fig.~\ref{fig:lambda_sensitivity}, the regularization intensity $\lambda$ controls the equilibrium between seen domain fitting and unseen domain generalization. A lower $\lambda$ favors $AO$ accuracy but limits transferability, while an excessively high $\lambda$ may over-restrict the representation space. Our results show that at $\lambda = 1.0$, the model achieves the optimal Harmonic Mean, effectively mitigating the Semantic Gap without collapsing on seen categories.}

\noindent{\textbf{Quantitative Analysis of the Semantic Gap.}}
{To further clarify the operational boundaries of SPA, we conduct a granular analysis on the MIT-States dataset by partitioning unseen test samples into three groups—Near, Moderate, and Isolated—based on their semantic similarity to the closest training-set neighbors. As summarized in Table~\ref{tab:q2_mit_isolation}, SPA's performance gain is positively correlated with semantic proximity, achieving a significant +4.93\% improvement in the "Near" group. Crucially, even in the "Isolated" group (Avg. Sim = 0.41), where semantic anchors are sparse, SPA maintains a consistent gain of +1.51\% over the baseline. This demonstrates that our Structure-Consistent Learning (SCL) objective provides effective global manifold regularization, ensuring a robust "performance floor" and preventing representation collapse even in semantically sparse regions.}

\begin{table}[tbp]
\centering
\scriptsize
\caption{{\textbf{Performance Gain vs. Semantic Isolation on MIT-States.}
Samples are partitioned into three groups based on the semantic similarity to their closest neighbors in the training set. 
The results reflect the $A^*O^*$ performance for CSP and CSP+SPA.}}
\label{tab:q2_mit_isolation}
\begin{tabularx}{\linewidth}{@{}l c c *{3}{>{\centering\arraybackslash}X}@{}}
\toprule
\textbf{Isolation Group} & \textbf{Ratio} & \textbf{Avg. Sim} & \textbf{CSP ($A^*O^*$)} & \textbf{+SPA ($A^*O^*$)} & \textbf{Gain ($\Delta$)} \\
\midrule
\rowcolor{green!5}
Near     & 50\% & 0.78 & 25.69 & 30.62 & \textbf{+4.93} \\
\rowcolor{yellow!5}
Moderate       & 40\% & 0.66 & 25.37 & 28.63 & \textbf{+3.26} \\
\rowcolor{red!5}
Isolated & 10\% & 0.41 & 25.27 & 26.78 & \textbf{+1.51} \\
\midrule
\rowcolor{gray!10}
\textbf{Total Average} & 100\% & 0.70 & 25.52 & 29.44 & \textbf{+3.92} \\
\bottomrule
\end{tabularx}
\end{table}

\subsection{Qualitative Results}
\label{sec_vis}

To complement our quantitative analysis and provide a deeper understanding of SPA’s behavior, we present a qualitative evaluation comparing our full model (CSP+SPA) against the baseline CSP on MIT-States and C-GQA datasets in Fig.~\ref{vis1}. This analysis helps reveal how SPA handles diverse compositions and highlights its strengths and limitations in open-vocabulary reasoning.

\noindent\textbf{Success Cases.}
The left panel of Fig.~\ref{vis1} demonstrates SPA's effectiveness in correcting the baseline's semantic errors, particularly on challenging open-vocabulary compositions. This success stems from SPA's ability to leverage learned structural relationships to infer the meaning of unseen primitives more precisely. For example, CSP confuses the unseen attribute "barren" with the semantically related but incorrect "weathered". SPA corrects this, demonstrating a finer-grained semantic understanding. This improvement stems from SPA’s ability to align the distributions of unseen primitives with semantically similar seen ones, enabling better generalization. 
Concurrently, SPA maintains high accuracy on seen compositions (\eg 'windblown desert'), demonstrating that SCL effectively regularizes for generalization without degrading seen-domain performance. 

\noindent\textbf{Failure Cases and Limitations.}
The right panel of Fig.~\ref{vis1} reveals the method's boundaries when there is a large semantic gap between the unseen concept and its closest seen neighbors. A deeper analysis shows that even in these failure cases, SPA often demonstrates more robust semantic reasoning than the baseline. Consider the $A^*O^*$ example with the ground truth "Intricate artwork". 
{The baseline CSP fails by predicting 'striped wallpaper', likely misled by superficial visual textures. Conversely, CSP+SPA predicts "Intricate sculpture". It correctly identifies the highly abstract, unseen attribute "Intricate" and recognizes a far more appropriate object category ("sculpture" \vs "artwork"). This suggests that our structural analogy mechanism captures higher-order semantic patterns that transcend simple pixel-level matching, providing a stable performance floor even when local semantic anchors are sparse. Ultimately, failures typically arise only when the semantic gap is exceptionally vast, causing the analogical reasoning of SAS to lack reliable concepts for knowledge transfer. This highlights the method's reliance on semantic proximity and points toward avenues for future work.}

\section{Conclusion}

In this paper, we explore how to improve compositional generalization in Open-Vocabulary Compositional Zero-Shot Learning (OV-CZSL) through prompt tuning methods. Specifically, we propose SPA, an effective plug-in method that aligns the structure of seen and unseen attributes and objects to enhance generalization. By leveraging the local structural relationships among semantically similar attributes and objects, SPA preserves meaningful structures during training through a Structure-aware Consistency Loss (SCL), and adapts unseen concepts at inference via a Structure-guided Adaptation Strategy (SAS). Extensive experiments demonstrate that SPA significantly improves performance in the open-vocabulary setting while maintaining strong results in the conventional CZSL scenario, highlighting its effectiveness and applicability.

\bibliographystyle{IEEEtranN}
\bibliography{my_references}


\clearpage 
\appendix 
\setcounter{section}{0}
\setcounter{figure}{0}
\setcounter{table}{0}
\renewcommand{\thesection}{S\arabic{section}}
\renewcommand{\thefigure}{S\arabic{figure}}
\renewcommand{\thetable}{S\arabic{table}}

\section*{Supplementary Material}

\section*{Overview}
This supplementary material provides additional experimental details for our SPA method. 
Specifically, we include: (1) a detailed breakdown of the \textbf{dataset splits} for the four OV-CZSL benchmarks; (2) detailed \textbf{ablation studies} on the hyperparameters $\lambda$ and $\tau$, which control the weight of the Structure-aware Consistency Loss (SCL) and the temperature for neighbor similarity, respectively; (3) a \textbf{comprehensive sensitivity analysis} of the neighbor count $K$, demonstrating its remarkable stability and scalability across multiple datasets with varying vocabulary densities, and (4) a \textbf{scalability analysis} of the Structure-guided Adaptation Strategy (SAS) module, highlighting computational bottlenecks and potential optimizations for large vocabulary settings. 
These results complement the analyses presented in the main paper and further demonstrate the robustness and efficiency of our approach.

\section*{Dataset Splits}
Table~\ref{datasplits} summarizes the dataset splits for the {four} OV-CZSL benchmarks. We follow a standard convention, where $A/A^*$ and $O/O^*$ denote seen/unseen attributes and objects, and the validation/test sets are further categorized into five compositional types: $AO$, $(AO)^*$, $A^*O$, $AO^*$, and $A^*O^*$. This allows for a systematic evaluation of generalization to unseen compositions.

\begin{table*}[hbp]
\caption{\textbf{Dataset splits for the three OV-CZSL benchmarks.} Notation: $A/A^*$ and $O/O^*$ denote seen/unseen attributes and objects. The val/test sets are broken down into five compositional types: seen attribute–seen object pairs ($AO$), novel compositions of seen attributes and objects ($(AO)^*$), unseen attribute–seen object pairs ($A^*O$), seen attribute–unseen object pairs ($AO^*$), and fully novel pairs ($A^*O^*$). This partitioning enables a comprehensive evaluation of generalization to novel compositions and unseen concepts.}
\centering
\begin{tabular}{c cccccc c}
\toprule
\multirow{2}{*}{Datasets} & \multicolumn{2}{c}{Attributes} & \multicolumn{2}{c}{Objects} & \multicolumn{1}{c}{Training Set} & \multicolumn{1}{c}{Validation Set} & \multicolumn{1}{c}{Test Set} \\
\cmidrule(lr){2-3} \cmidrule(lr){4-5} \cmidrule(lr){6-6} \cmidrule(lr){7-7} \cmidrule(lr){8-8}
 & $A$ & $A^*$ & $O$ & $O^*$ & $AO$ & $AO$/$(AO)^*/A^*O/AO^*/A^*O^*$ & $AO$/$(AO)^*/A^*O/AO^*/A^*O^*$ \\
\midrule
MIT-States~\cite{isola2015discovering} & 84 & 31 & 182 & 63 & 955 &  236 / 105 / 126 / 177 / 44 & 289 / 130 / 157 / 218 / 50 \\
C-GQA~\cite{mancini2022learning} & 311 & 102 & 504 & 170 & 4094 &  1012 / 447 / 525 / 517 / 147 & 1239 / 542 / 664 / 655 / 176 \\
VAW-CZSL~\cite{pham2021learning} & 330 & 135 & 406 & 110 & 7142 & 1767 / 803 / 1420 / 1253 / 412 &  2161 / 982 / 1737 / 1532 / 504 \\
UT-Zappos~\cite{DBLP:conf/cvpr/YuG14} & 12 & 4 & 9 & 3 & 49 & 8 / 9 / 6 / 8 / 2 &  8 / 9 / 7 / 8 / 2 \\
\bottomrule
\end{tabular}
\label{datasplits}
\end{table*}

\section*{Ablation study on Hyperparameters}
\noindent{\textbf{Effect of hyper-parameter $\lambda$.}} 
We analyze the impact of the SCL loss weight $\lambda$, which balances the primary classification objective with our structural consistency regularizer. Tab.~\ref{tab:ablation_lamda} shows that performance is stable across a wide range of $\lambda$ values, underscoring the robustness of our method. The model achieves the best overall performance at $\lambda$=1.00, reaching the highest HM (27.80) and a top $A^*O^*$ (29.44). While a very large weight ($\lambda$=10.00) can slightly improve the $A^*O^*$ metric further, it comes at the cost of the overall HM, suggesting that an overly strong structural constraint can begin to interfere with the classification task. Thus, we identify $\lambda$=1.00 as the optimal value.

\begin{table}[H]
\centering
\small
\caption{\textbf{Ablation study of the SCL weight $\lambda$ on MIT-States}, showing that performance is stable across a wide range of values, with $\lambda=1.0$ achieving the best overall performance.}
\label{tab:ablation_lamda}
\begin{tabular}{@{} l | S[table-format=2.2] S[table-format=2.2] S[table-format=2.2] S[table-format=2.2] S[table-format=2.2] @{}}
\toprule
\multirow{2}{*}{Weight $\lambda$} & \multicolumn{5}{c}{MIT-States} \\
& {HM$\uparrow$} & {AUC$\uparrow$} & {$A^*O$$\uparrow$} & {$AO^*$$\uparrow$} & {$A^*O^*$$\uparrow$} \\
\midrule
$\lambda = 0.05$ & 27.52 & 10.58 & 23.15 & 26.66 & 28.74 \\
$\lambda = 0.10$ & 27.49 & 10.58 & 23.20 & 26.58 & 28.93 \\
$\lambda = 0.50$ & 27.52 & 10.58 & 23.03 & 26.69 & 28.68 \\
\rowcolor{blue!6}
$\lambda = 1.00$ & \textbf{27.80} & \textbf{10.59} & \textbf{26.52} & 27.38 & 29.44 \\
$\lambda = 5.00$ & 27.49 & 10.52 & 23.01 & 26.69 & 29.00 \\
$\lambda = 10.00$ & 27.47 & 10.45 & 25.34 & \textbf{27.78} & \textbf{29.75} \\
\bottomrule
\end{tabular}
\vspace{-0.5em}
\end{table}

\noindent{\textbf{Effect of hyper-parameter $\tau$.}} 
The temperature $\tau$ controls the sharpness of the neighbor similarity distribution in SCL. We investigate its effect in Tab.~\ref{tab:ablation_tau}. The results indicate that model performance is sensitive to this parameter, with a clear optimal value at $\tau$=0.10. A lower temperature (\eg 0.01) makes the distribution too peaky and may cause the model to rely on a few potentially noisy neighbors, while a higher temperature (\eg 0.20) softens the distribution excessively, giving undue weight to less relevant concepts. A value of $\tau$=0.10 appears to effectively focus the model on the most semantically relevant neighbors, leading to the best performance across all key metrics.

\begin{table}[ht]
\centering
\caption{\textbf{Ablation study of the temperature $\tau$ in SPA on MIT-States}, illustrating that performance peaks at $\tau=0.10$, which provides the best HM and open-vocabulary scores.}
\label{tab:ablation_tau}
\begin{tabular}{c | S[table-format=2.2] S[table-format=2.2] S[table-format=2.2] S[table-format=2.2] S[table-format=2.2]}
\toprule
\multirow{2}{*}{Temperature $\boldsymbol{\tau}$} & \multicolumn{5}{c}{MIT-States} \\
& {HM$\uparrow$} & {AUC$\uparrow$} & {$A^*O$$\uparrow$} & {$AO^*$$\uparrow$} & {$A^*O^*$$\uparrow$} \\
\midrule
$\tau = 0.01$ & 27.23 & 10.25 & 26.22 & 26.24 & 28.52 \\
$\tau = 0.05$ & 27.38 & 10.42 & 25.55 & 26.69 & 28.50 \\
$\tau = 0.07$ & 27.47 & 10.46 & 25.53 & 26.77 & 29.31 \\
\rowcolor{blue!6} 
$\tau = 0.10$ & \textbf{27.80} & \textbf{10.59} & \textbf{26.52} & \textbf{27.38} & \textbf{29.44} \\
$\tau = 0.20$ & 27.52 & 10.46 & 23.15 & 26.72 & 28.49 \\
\bottomrule
\end{tabular}
\end{table}

\noindent{\textbf{Effect of hyper-parameter $K$ on C-GQA and VAW-CZSL.}}
To further investigate the sensitivity and scalability of the number of neighbors K, we extend our ablation study to the larger-scale C-GQA and VAW-CZSL datasets. As shown in Table~\ref{tab:ablation_k_cgqa} and Table~\ref{tab:ablation_k_vaw}, SPA exhibits remarkable stability across varying vocabulary scales. In both benchmarks, the performance consistently peaks at K=5, which achieves the highest Harmonic Mean (HM) and significant gains in unseen-split metrics such as $A^*O^*$. These results confirm the principle of intrinsic semantic locality, which suggests that the meaningful structural information of a concept is contained within a limited set of its closest semantic neighbors. While the total vocabulary size increases, the number of truly relevant semantic anchors remains relatively constant. Consequently, $K=5$ serves as a robust sweet spot that effectively captures the core semantic essence without introducing neighbor-induced interference, proving that SPA is not strictly dependent on global vocabulary density.

\begin{table}[H]
\centering
\small
\caption{\textbf{Ablation study on the effect of the number of neighbors ($K$) in SPA.} We report key performance metrics on C-GQA, showing that a moderate choice of $K=5$ achieves the best trade-off across HM and open-vocabulary metrics.}
\label{tab:ablation_k_cgqa}
\begin{tabular}{@{} c | S[table-format=2.2] S[table-format=2.2] S[table-format=2.2] S[table-format=2.2] S[table-format=2.2] @{}}
\toprule
\multirow{2}{*}{\textbf{Neighbors $K$}} & \multicolumn{5}{c}{C-GQA} \\
& {HM$\uparrow$} & {AUC$\uparrow$} & {\textbf{$A^*O$$\uparrow$}} & {\textbf{$AO^*$$\uparrow$}} & {\textbf{$A^*O^*$$\uparrow$}} \\
\midrule
$K = 0$ (Baseline) & 15.19 & 3.54 & 16.49 & 7.65 & 9.28 \\
$K = 1$ & 15.27 & 3.66 & 16.14 & 7.15 & 5.38 \\
$K = 3$ & 15.90 & 3.87 & 16.90 & 7.78 & 7.11 \\
\rowcolor{blue!6}
$K = 5$ & \textbf{16.70} & \textbf{4.07} & \textbf{18.98} & \textbf{7.12} & \textbf{10.67} \\
$K = 7$ & 16.65 & 4.03 & 18.45 & 7.34 & 10.44 \\
$K = 10$ & 16.57 & 3.95 & 18.71 & 7.54 & 10.37 \\
\bottomrule
\end{tabular}
\end{table}

\begin{table}[H]
\centering
\small
\caption{\textbf{Ablation study on the effect of the number of neighbors ($K$) in SPA.} We report key performance metrics on VAW-CZSL, showing that a moderate choice of $K=5$ achieves the best trade-off across HM and open-vocabulary metrics.}
\label{tab:ablation_k_vaw}
\begin{tabular}{@{} c | S[table-format=2.2] S[table-format=2.2] S[table-format=2.2] S[table-format=2.2] S[table-format=2.2] @{}}
\toprule
\multirow{2}{*}{\textbf{Neighbors $K$}} & \multicolumn{5}{c}{VAW-CZSL} \\
& {HM$\uparrow$} & {AUC$\uparrow$} & {\textbf{$A^*O$$\uparrow$}} & {\textbf{$AO^*$$\uparrow$}} & {\textbf{$A^*O^*$$\uparrow$}} \\
\midrule
$K = 0$ (Baseline) & 13.26 & 2.71 & 3.21 & 5.55 & 5.59 \\
$K = 1$ & 12.79 & 2.56 & 2.67 & 5.30 & 4.39 \\
$K = 3$ & 13.05 & 2.73 & 2.85 & 4.98 & 6.14 \\
\rowcolor{blue!6}
$K = 5$ & \textbf{13.92} & 2.42 & \textbf{4.36} & \textbf{6.40} & \textbf{6.70} \\
$K = 7$ & 13.86 & 2.37 & 4.34 & 6.34 & 6.67 \\
$K = 10$ & 13.91 & 2.39 & 4.37 & 6.28 & 6.65 \\
\bottomrule
\end{tabular}
\end{table}

\section*{Scalability Analysis}
The scalability of the SAS module may be a concern for OV-CZSL tasks with extremely large vocabularies. The overhead of this module can be broken down into two main components: (1) a k-Nearest Neighbor (k-NN) search to find relevant seen concepts, and (2) the subsequent adaptation and aggregation computation. To quantify the bottleneck, we profiled the inference time of these components, with results shown in Tab.~\ref{tab:cost_sas}. The analysis clearly indicates that the brute-force k-NN search is the dominant factor, accounting for over 80\% of the module's computational overhead. The k-NN search can be readily replaced by more efficient Approximate Nearest Neighbor (ANN) algorithms (\eg Faiss, ScaNN). This standard practice reduces the search complexity from linear to sub-linear, which in turn enhances SPA's applicability to massive-scale vocabularies.

\begin{table}[ht]
\centering
\caption{Breakdown of the \textbf{inference time} overhead of the SAS module.}
\label{tab:cost_sas}
\begin{tabular}{@{}lc@{}}
\toprule
Component in SAS   & Share of Inference Time (\%) \\ \midrule
k-NN Search                 & \textbf{80.2\%}                      \\
Adaptation \& Aggregation & 19.8\%                      \\ \bottomrule
\end{tabular}
\end{table}

\vfill

\end{document}